\pdfoutput=1
\documentclass[sigconf]{acmart}
\AtBeginDocument{%
  }

\setcopyright{acmlicensed}
\copyrightyear{2018}
\acmYear{2018}
\acmDOI{XXXXXXX.XXXXXXX}
\acmConference[Conference acronym 'XX]{Make sure to enter the correct
  conference title from your rights confirmation email}{June 03--05,
  2018}{Woodstock, NY}
\acmISBN{978-1-4503-XXXX-X/2018/06}

\usepackage[T1]{fontenc}
\usepackage{newunicodechar}
\newunicodechar{≤}{\ensuremath{\le}}
\usepackage{graphicx}
\usepackage{transparent}
\usepackage{soul}

\usepackage{amsmath, amssymb} 
\usepackage[most]{tcolorbox}
\tcbuselibrary{breakable, skins}
\usepackage{booktabs}
\usepackage{multirow}
\usepackage[table]{xcolor}
\usepackage{adjustbox}
\usepackage{amssymb}
\usepackage{pifont}
\usepackage{xcolor}
\usepackage{amssymb}
\newcommand{\cmark}{\textcolor{green!60!black}{\checkmark}}
\newcommand{\xmark}{\textcolor{red}{\ding{55}}}

\tcbuselibrary{listings, breakable}

\begin{document}

\title{Reflective Personalization Optimization: A Post-hoc Rewriting Framework for Black-Box Large Language Models}

\author{Teqi Hao}
\authornote{Both authors contributed equally to this work.}
\affiliation{%
  \institution{School of Electronic and Electrical Engineering, Shanghai University of Engineering Science}
  \city{Shanghai}
  \state{Shanghai}
  \country{China}
}

\author{Xiaoyu Tan}
\authornotemark[1]
\affiliation{%
  \institution{Tencent Youtu Lab}
  \city{Shanghai}
  \state{Shanghai}
  \country{China}
}

\author{Shaojie Shi}
\affiliation{%
  \institution{Artificial Intelligence Innovation and Incubation Institute, Fudan University}
  \city{Shanghai}
  \state{Shanghai}
  \country{China}
}

\author{Yinghui Xu}
\affiliation{%
  \institution{Artificial Intelligence Innovation and Incubation Institute, Fudan University}
  \city{Shanghai}
  \state{Shanghai}
  \country{China}
}

\author{Xihe Qiu}
\authornote{Corresponding author.}
\affiliation{%
  \institution{School of Electronic and Electrical Engineering, Shanghai University of Engineering Science}
  \city{Shanghai}
  \state{Shanghai}
  \country{China}
}
\email{qiuxihe1993@gmail.com}


\begin{abstract}
The personalization of black-box large language models (LLMs) is a critical yet challenging task. Existing approaches predominantly rely on context injection, where user history is embedded into the prompt to directly guide the generation process. However, this single-step paradigm imposes a dual burden on the model: generating accurate content while simultaneously aligning with user-specific styles. This often results in a trade-off that compromises output quality and limits precise control. To address this fundamental tension, we propose Reflective Personalization Optimization (RPO), a novel framework that redefines the personalization paradigm by decoupling content generation from alignment. RPO operates in two distinct stages: first, a base model generates a high-quality, generic response; then, an external reflection module explicitly rewrites this output to align with the user's preferences. This reflection module is trained using a two-stage process. Initially, supervised fine-tuning is employed on structured rewriting trajectories to establish a core personalized reasoning policy that models the transformation from generic to user-aligned responses. Subsequently, reinforcement learning is applied to further refine and enhance the quality of the personalized outputs. Comprehensive experiments on the LaMP benchmark demonstrate that RPO, by decoupling content generation from personalization, significantly outperforms state-of-the-art baselines. These findings underscore the superiority of explicit response shaping over implicit context injection. Moreover, RPO introduces an efficient, model-agnostic personalization layer that can be seamlessly integrated with any underlying base model, paving the way for a new and effective direction in user-centric generation scenarios. \textbf{The code will be open-sourced due to acceptance.}
\end{abstract}



\maketitle

\section{Introduction}

Pre-trained Large Language Models (LLMs) have demonstrated remarkable capabilities across a wide range of Natural Language Processing (NLP) tasks, including recommender systems, dialogue agents, information retrieval, and text summarization\cite{liao2025avoid, lin2025rec, shim2025tooldial, zhang2024escot, maiya2025onprem, shi2025direct, deng2024text}. Their broad applicability stems from strong cross-domain generalization, enabling competitive performance even in zero-shot and few-shot settings\cite{sanh2021multitask, zhang2025nature, brown2020language, vu2021strata}. However, while most LLMs generate fluent and coherent text, they typically follow a generic "one-size-fits-all" generation paradigm, which struggles to capture user-specific nuances\cite{chen2024large, christakopoulou2023large, tan2023user}. In personalized scenarios—such as tailored dialogues, writing assistance, or content recommendation—this generic behavior often results in outputs that are structurally sound yet semantically misaligned with user expectations\cite{li2024personal, kang2023llms}.

\begin{figure}[!t]
  \includegraphics[width=\columnwidth]{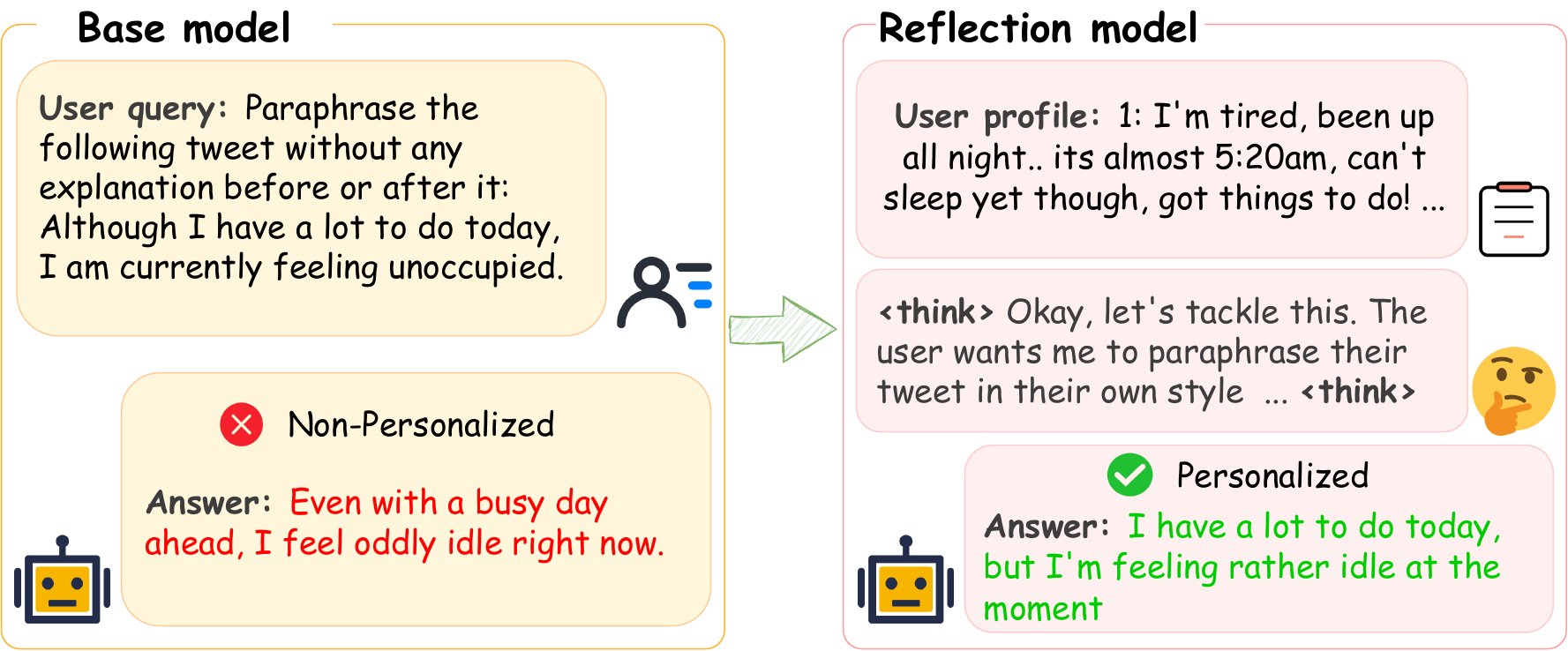}
  \caption{This diagram illustrates the workflow of RPO. The system first utilizes a Base model to generate an initial response without personalized information based on the user query. Subsequently, the reflection module integrates user profile information and refines the initial response through reasoning enhancement, ultimately generating output that aligns with the user's personalized needs.}
  \label{fig:abstract}
  \vspace{-3mm}
\end{figure}

Personalization has thus emerged as a key challenge, particularly for proprietary, black-box base models that are too large for conventional fine-tuning. The prevailing strategy for these models is context injection, where user history is prepended to the prompt to guide a direct generation process. This monolithic approach forces the LLM to simultaneously tackle two often conflicting objectives: generating factually accurate content and aligning with nuanced user styles, creating an inherent tension that leads to outputs that can be either stylistically weak or factually compromised~\cite{han2023persona, zhao2023unimc}. More fundamentally, these methods treat personalization as a black-box conditioning task, failing to explicitly model the latent reasoning policy that governs a user's stylistic choices. As a result, personalization becomes uncontrollable and inconsistent, with the model merely guessing the user's intent from raw data rather than learning their underlying decision-making process~\cite{han2023persona, zhao2023unimc}.

Moreover, these methods provide only implicit, uncontrollable guidance, and are often limited by context length~\cite{borgeaud2022improving} and prone to inconsistency across sessions~\cite{han2023persona, zhao2023unimc}. While Parameter-Efficient Fine-Tuning (PEFT) techniques—such as LoRA, adapters, and prompt tuning~\cite{hu2022lora, pfeiffer2020adapterfusion, lester2021power}—offer an alternative, they are infeasible for black-box models as they require access to internal parameters. This leaves a critical gap: \textbf{how to achieve robust, controllable personalization for inaccessible models without compromising content integrity?}

\begin{table*}[t]
\centering
\small
\caption{Comparison of personalization approaches.}
\renewcommand{\arraystretch}{1.2}
\setlength{\tabcolsep}{6pt}
\begin{tabular}{lccc}
\toprule
\textbf{Comparison Aspect} & \textbf{Retrieval-Augmented} & \textbf{PEFT Methods} & \textbf{RPO (Ours)} \\
\midrule
Personalization Strategy        & In-context Prompting & Parameter Tuning & \textbf{Post-hoc Rewriting} \\
Requires base model tuning      & \xmark & \cmark & \textbf{\xmark} \\
User profile utilization        & Weak & Moderate & \textbf{Strong} \\
Generalization to new users     & Limited & Moderate & \textbf{High} \\
Base model agnostic             & \cmark & \xmark & \textbf{\cmark} \\
Interpretability                & \xmark & \xmark & \textbf{\cmark} \\
\bottomrule
\end{tabular}
\label{tab:rpo-comparison}
\end{table*}

To address the challenges of personalization in language models, we propose a paradigm shift from direct generation to a post-hoc response shaping approach. This method emphasizes not just rewriting, but doing so in a principled and interpretable manner. At the core of our approach lies the concept of \textbf{Structured Rewriting Trajectories}, a novel data representation that externalizes the process of transforming a generic response into one that aligns with a user's preferences. Structured rewriting trajectories provide a rich, interpretable reasoning chain that explicitly links user profile attributes to specific rewriting actions, making the user's latent reasoning policy observable and learnable~\cite{madaan2023self, li2023teach}. Building on this foundation, we introduce \textbf{Reflective Personalization Optimization (RPO)}, a framework that formalizes a generate-then-rewrite pipeline. RPO first leverages a powerful black-box LLM to generate high-quality, generic responses, preserving the semantic integrity of the output. It then applies an external reflection module to adapt these responses to the user’s linguistic style and intent. This modular design ensures compatibility with any black-box LLM~\cite{dathathri2019plug}, offering flexibility, modularity, and control without altering the base model's parameters.

The RPO framework consists of two key components: the \textbf{User Retriever} and the \textbf{Reflection Optimizer}. The User Retriever dynamically selects relevant user history based on the input, providing personalized signals for the rewriting process. The Reflection optimizer then refines the initial response iteratively through a multi-step reasoning chain, ensuring alignment with user preferences. To train this system effectively, we adopt a two-stage pipeline. The first stage involves supervised fine-tuning on structured rewriting trajectories, which establishes a strong foundational reasoning capability~\cite{ouyang2022training, bai2022training, li2024personalized}. In the second stage, reinforcement learning is employed, guided by a progressive multi-context curriculum. This curriculum gradually exposes the model to increasingly complex user contexts, enabling it to distill stable preferences from diverse and potentially noisy historical data. This two-stage approach ensures robustness and generalization, making RPO well-suited for real-world personalization tasks~\cite{qiuwisdom, chen2024u}.

We evaluate RPO on the LaMP personalization benchmark\cite{salemi2024optimization, kumar2024longlamp}, covering both classification and generation tasks. Experimental results demonstrate that RPO's decoupled approach consistently and significantly outperforms existing prompt-based, retrieval-augmented, and parameter-efficient methods across all settings\cite{salemi2024optimization}, while maintaining high modularity and broad model compatibility.

In summary, our contributions are as follows:
\begin{itemize}
    \item We identify the inherent conflict between content generation and stylistic reasoning in direct-generation personalization models and propose a decoupled generate-then-rewrite framework as a robust solution.
    \item We propose the RPO framework, which operationalizes this paradigm. Its core innovation is the structured rewriting trajectory, which transforms personalization from a black-box task into a learnable and interpretable reasoning process.
    \item We design a novel training pipeline featuring a progressive multi-context curriculum that significantly enhances the model's ability to generalize from diverse and noisy user histories, boosting its real-world applicability.
\end{itemize}

\section{Related work}

\subsection{Personalization via Direct Generation}

As LLMs continue to serve as general-purpose engines across NLP tasks\cite{brown2020language}, personalization has emerged as a critical frontier. Recent efforts aim to condition generation on user-specific information\cite{mireshghallah2021useridentifier}, moving beyond coarse task-level generalization toward fine-grained, user-aligned outputs.

The predominant approach for black-box models can be framed as personalization via direct generation. These methods attempt to produce a user-aligned output in a single forward pass, mainly through context injection. Retrieval-based approaches like RAG insert selected user history into the context window\cite{salemi2024optimization}, offering efficiency and model-agnosticity. However, this forces the model to resolve the "content-style conflict" implicitly. The model must infer how to use the provided context for personalization while simultaneously constructing the core response.

Other context injection variants include prompt and embedding tuning\cite{lester2021power, li2021prefix}, which are lightweight but often fall short in controllability and alignment granularity. An alternative "direct generation" strategy is parameter adaptation. PEFT techniques, such as LoRA\cite{hu2022lora} and adapters\cite{pfeiffer2020adapterfusion}, adapt lightweight modules for users. While powerful, they are inapplicable to black-box models and still train the model for a monolithic generation task, grappling with the same content-style trade-offs, albeit at the parameter level.

\subsection{Personalization via Response Refinement}

An alternative paradigm is emerging that decouples generation from refinement, aligning with our proposed approach. The concept of iterative refinement or self-correction, where a model revises its own output, has shown significant promise for improving general output quality dimensions like factuality and logical coherence\cite{madaan2023self}.

However, the systematic application of this "rewriting" paradigm to the complex, multi-faceted problem of user personalization remains largely unexplored. Personalization requires alignment with subjective stylistic nuances and historical preferences, not just objective quality metrics. Crucially, most prior work assumes static personalization and lacks mechanisms for such dynamic, targeted refinement\cite{wang2024ai}. This is the gap our work addresses. RPO is the first framework to formalize and build a dedicated training pipeline around this idea for personalization. We reframe the task not as direct generation, but as a post-hoc response shaping task, where an initial generic response is explicitly rewritten to achieve fine-grained user alignment.

\subsection{Reinforcement Learning for Alignment}

RL is gradually emerging as a core paradigm for aligning LLMs with individual user preferences\cite{li2024personalized}. While large-scale supervised pretraining endows LLMs with strong capabilities, these models still struggle to capture fine-grained user-specific attributes\cite{liu2025survey} such as personal style and tone. RL introduces an iterative optimization process guided by external reward signals, offering a more flexible and goal-driven alignment mechanism.

Currently, reinforcement learning from human feedback (RLHF) represents the dominant approach\cite{stiennon2020learning}, with successful applications in systems such as InstructGPT\cite{ouyang2022training} and Claude\cite{bai2022training}. However, RLHF’s reliance on manual preference data imposes scalability and cost limitations\cite{lee2023rlaif}. To address this, recent methods like RLAIF\cite{bai2022training} and RLVR\cite{luong2024reft} incorporate automatically generated reward signals, broadening the applicability of RL.

At the same time, methods such as DPO\cite{rafailov2023direct}, GRPO\cite{ramesh2024group}, and REINFORCE++\cite{hu2025reinforce++} mitigate the reliance on high-variance value functions, enhancing RL's feasibility. For example, the DeepSeek-R1 model\cite{guo2025deepseek} uses a multi-stage GRPO strategy for strong reasoning. These advancements make RL an ideal tool for the second stage of our RPO framework. After SFT provides the reflection module with a strong baseline rewriting policy, RL offers a powerful mechanism to further optimize this policy specifically for the response shaping task. It allows the model to explore the vast space of possible rewrites and learn to better balance multiple, sometimes competing, personalization goals (e.g., matching tone vs. incorporating specific keywords) based on a reward signal that reflects final output quality, thereby enhancing the reflection module's ability to map user characteristics to model generation behaviors.

\begin{figure*}[ht]
    \centering
    \adjustbox{margin=0cm 0cm 0cm 0cm}{\includegraphics[width=\textwidth]{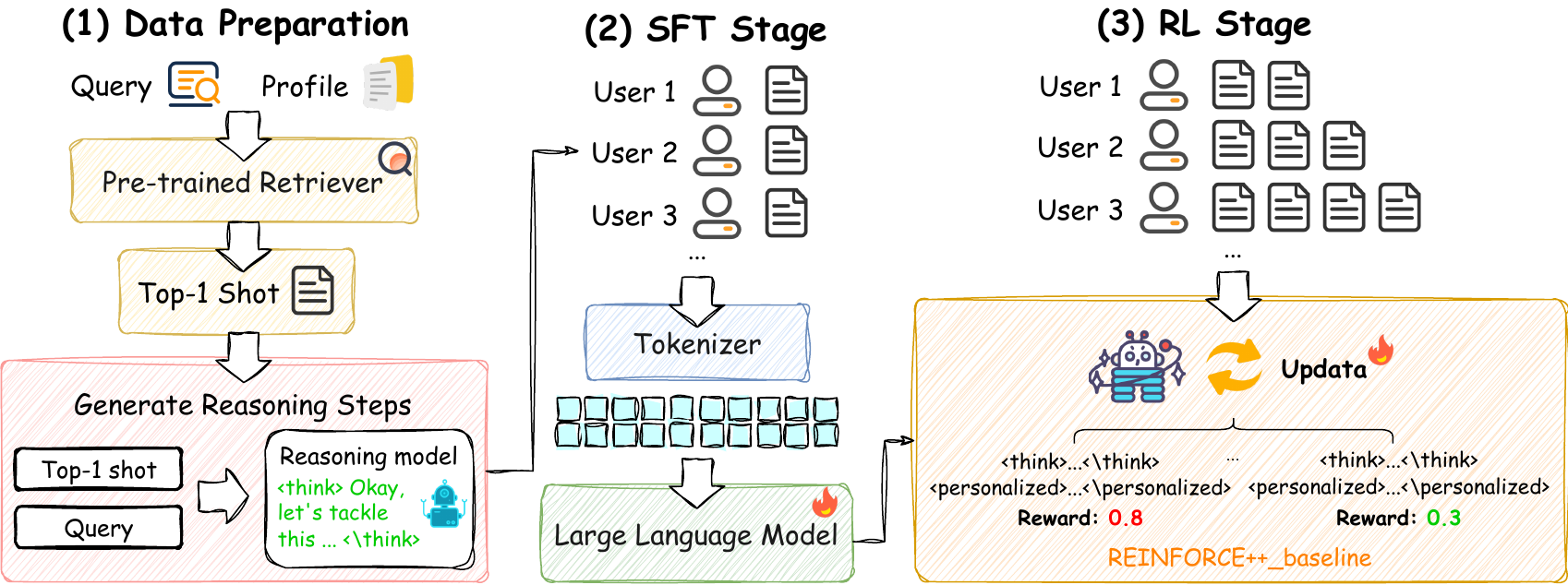} }
    \caption{This figure illustrates the training pipeline of the RPO framework, which encompasses three pivotal stages: (1) \textbf{Data Preparation}: The pipeline begins by generating a corpus of Structured Rewriting Trajectories. For each training instance, a powerful teacher model is prompted with a generic response and a user's historical example to externalize the latent reasoning process behind personalization. (2) \textbf{SFT Stage}: The reflection model is then trained on these trajectories via supervised fine-tuning. This stage aims to instill a foundational personalized reasoning policy by having the model learn the explicit, step-by-step logic demonstrated in the trajectories. (3) \textbf{RL Stage}: Finally, the model's policy is refined using reinforcement learning. This stage features a key innovation: a progressive multi-context curriculum. The model is trained on a varying number of user profile examples (from 2 to 6) to enhance its ability to generalize from diverse and noisy user histories. This process is optimized using the \texttt{REINFORCE++\_baseline} method.}
    \label{fig:framework}
\end{figure*}

\section{Method}
\label{sec:method}

This section details the Reflective Personalization Optimization (RPO) framework, our proposed solution to operationalize the decoupled alignment paradigm. As illustrated in Figure~\ref{fig:framework}, RPO decomposes personalization into two distinct stages: (1) initial generation of a high-quality, generic response, and (2) subsequent, reasoning-driven response shaping via an external reflection module.

\subsection{Framework Overview and Problem Formulation}
\label{sec:overview}

The RPO framework is architecturally decomposed into two key components:

\begin{itemize}
    \item \textbf{The Base Model}, denoted as $M_{\text{base}}$, can be any pre-trained LLM, treated strictly as a black box. Given a user query $q$, its sole role is to produce a general, non-personalized response $A_{\text{base}}$, such that $A_{\text{base}} = \mathcal{M}_{\text{base}}(q)$. By offloading this task to a powerful base model, we ensure the semantic integrity of the core content.

    \item \textbf{The Reflection Model}, denoted as $M_{\text{reflect}}$, is a separate, fully accessible language model that performs the core response shaping task. In our framework, this is a powerful, instruction-tuned language model. For each user, we maintain a profile of historical interactions $U=\{p_{1},...,p_{N}\}$. A retriever first selects a subset of relevant history, which we term the personalized context $P_{\text{rel}} \subseteq U$.
\end{itemize}

The core research challenge then becomes: how to train $M_{\text{reflect}}$ to effectively transform the generic response $A_{\text{base}}$ into a user-aligned final answer $A_{\text{final}}$, based on the signals from the personalized context $P_{\text{rel}}$. This transformation is defined as the policy of the reflection model:
\begin{equation}
    A_{\text{final}} = \mathcal{M}_{\text{reflect}}(q, A_{\text{base}}, P_{\text{rel}})
\end{equation}
The key challenge lies not in merely identifying the optimal personalized context $P_{\text{rel}}$, but in training $M_{\text{reflect}}$ to learn the complex \textit{policy} of personalized rewriting, even from scarce or minimal user history.

\subsection{Modeling Personalized Reasoning via Structured Rewriting Trajectories}
\label{sec:trajectories}

As established in the introduction, the cornerstone of RPO is its ability to learn a user's \textit{latent reasoning policy}. To achieve this, we must first make this policy observable and machine-readable. We introduce a principled pipeline for generating structured rewriting trajectories. Each trajectory is a data instance showing the explicit reasoning path \cite{wei2022chain} that transforms a generic response into a user-aligned one. For specific cases, see Appendix \ref{app:prompt}.

The core of our data generation pipeline is a modular prompt template that guides the teacher LLM (gpt-4.1-mini-2025-04-14) to produce these trajectories. The template provides a starting point (the generic response $A_{\text{base}}$) and the conditioning evidence (a selected user profile example $p^{*}$), converting an unconstrained generation task into a more tractable, constrained rewriting problem. This structure compels the teacher model to perform a targeted differential analysis, identifying salient features from the user profile that explicitly account for the delta between the generic and personalized responses. Consequently, the elicited thinking process correlates specific user attributes with concrete rewriting actions, making the user's latent policy directly learnable.

Our supervised dataset, denoted as $\mathcal{D}_{\text{SFT}}$, is constructed by executing this pipeline at scale. For each ground-truth instance $(q, A_{\text{final}}^{\text{GT}})$, we generate $A_{\text{base}}$ and then prompt the teacher model to produce the full structured rewriting trajectory. We apply strict consistency and brevity checks to filter the dataset, ensuring a high-quality corpus for training our reflection model.

\subsection{Two-Stage Training of the Reflection Module}
\label{sec:training}

The training of $\mathcal{M}_{\text{reflect}}$ follows a two-stage paradigm designed to first instill foundational rewriting skills and then refine them for optimal performance.

\noindent \textbf{Step 1. SFT stage.}
The primary objective of the SFT stage is to teach the reflection model the fundamental patterns of personalized rewriting using our constructed dataset $\mathcal{D}_{\text{SFT}}$. This step grounds the model in the explicit, structured logic of the rewriting trajectories. For each training instance, which includes the user query $q$, the base response $A_{\text{base}}$, a single top-retrieved user example $p^{*}$, and the ground-truth answer $A^{\text{GT}}_{\text{final}}$, we minimize the sequence cross-entropy loss $\mathcal{L}_{\text{SFT}}$:
\begin{equation}
\label{eq:sft}
\small
\mathcal{L}_{\text{SFT}}
= - \sum_{t=1}^{T}
    \log P_{\theta}\!\left(
        A^{\text{GT}}_{\text{final},t}
        \mid q, A_{\text{base}}, p^{*},
        A^{\text{GT}}_{\text{final},<t}
    \right),
\end{equation}
where $\theta$ are the parameters of $\mathcal{M}_{\text{reflect}}$ and $T$ is the target sequence length.

\noindent \textbf{Step 2. RL stage.}
To further refine the SFT policy, denoted as $\pi_{\text{SFT}}$, and enhance generalization, we employ reinforcement learning. We cast generation as a Markov Decision Process (MDP), where the policy $\pi_{\theta}$ is optimized to maximize the expected return. For a generated answer $y$, the reward at the final step is:
\begin{equation}
\small
r_t
= \mathbb{I}(t = T)\,
  r\!\left(q, A_{\text{base}}, P_{\text{rel}}, y\right)
  - \beta\,\text{KL}_t,
\end{equation}
with the KL-divergence term $\text{KL}_t = \text{KL}\!\left[ \pi_{\theta}(\cdot \mid S_t)\; \big\|\; \pi_{\text{SFT}}(\cdot \mid S_t) \right]$ regularizing the policy update, controlled by $\beta$. The task reward $r(\cdot)$ is a composite score based on task-specific metrics (e.g., ROUGE, MAE, Acc/F1). We use the REINFORCE++\_baseline estimator for stable gradient computation.

\noindent \textbf{Progressive Multi-Context Generalization.} A key innovation in our RL stage is a progressive multi-context curriculum, designed to address the variable density and relevance of real-world user histories. To build a robust model, we eschew training on a fixed number of historical examples. Instead, the number of profile shots, $k$ (i.e., the size of $P_{\text{rel}}$), is dynamically determined by a curriculum scheduling function, $\mathcal{K}(e)$, which evolves with the training epoch $e$. Specifically, the policy $\pi_{\theta}$ is optimized to maximize an expected return where the complexity of the context distribution itself increases over time. We implement this through a simple yet effective stepwise schedule:
\begin{equation*}
    k = \mathcal{K}(e) = \min(k_{\text{max}}, k_{\text{min}} + \lfloor \frac{e-1}{E_{\text{step}}} \rfloor)
\end{equation*}
where we set a minimum of $k_{\text{min}}=2$ and a maximum of $k_{\text{max}}=6$ shots, and $E_{\text{step}}$ represents the number of epochs per step increase.

This curriculum learning approach compels the model to first capture a user's core style from a few, highly salient examples (when $k$ is small), before learning to synthesize preferences from more complex and potentially noisy contexts (as $k$ grows). By systematically increasing the difficulty of the context-conditioning task, this strategy significantly enhances the generalization capabilities of the policy and mitigates "style drift" caused by irrelevant information.

\subsection{Inference Workflow}
\label{sec:inference}

Given a trained RPO system, generating a personalized response for a new user query $q$ follows a three-step process:

\begin{enumerate}
    \item \textbf{Initial Response Generation:} The query $q$ is sent to the base model $M_{\text{base}}$, which produces the generic response $A_{\text{base}}$.

    \item \textbf{Personalized Context Retrieval:} We form a richer retrieval query by concatenating the user query and the initial response, $q \oplus A_{\text{base}}$. This is used to retrieve the top-$k$ most relevant context fragments from the user's historical profile $U$, forming the personalized context $P_{\text{rel}}$.

    \item \textbf{Reflective Refinement:} Finally, the trained reflection model $M_{\text{reflect}}$ receives the query $q$, the generic response $A_{\text{base}}$, and the personalized context $P_{\text{rel}}$. It then performs the final rewriting task to generate the personalized answer $A_{\text{final}}$.
\end{enumerate}

\begin{table*}[t]
  \centering
  \normalsize
  \setlength{\tabcolsep}{3.2pt} 
  \renewcommand{\arraystretch}{1.1}
  \caption{Performance of the RPO method on the LaMP benchmark.}
  \label{tab:lamp_results_progressive}
  \begin{tabular}{l cc cc cc cc cc cc cc cc}
    \toprule
    \multirow{3}{*}{\textbf{Method}} & \multicolumn{4}{c}{\textbf{LaMP-2}} & \multicolumn{4}{c}{\textbf{LaMP-3}} & \multicolumn{4}{c}{\textbf{LaMP-5}} & \multicolumn{4}{c}{\textbf{LaMP-7}} \\
    \cmidrule(lr){2-5} \cmidrule(lr){6-9} \cmidrule(lr){10-13} \cmidrule(lr){14-17}
    & \multicolumn{2}{c}{User} & \multicolumn{2}{c}{Time} & \multicolumn{2}{c}{User} & \multicolumn{2}{c}{Time} & \multicolumn{2}{c}{User} & \multicolumn{2}{c}{Time} & \multicolumn{2}{c}{User} & \multicolumn{2}{c}{Time} \\
    \cmidrule(lr){2-3} \cmidrule(lr){4-5} \cmidrule(lr){6-7} \cmidrule(lr){8-9} \cmidrule(lr){10-11} \cmidrule(lr){12-13} \cmidrule(lr){14-15} \cmidrule(lr){16-17}
    & Acc. $\uparrow$ & F1 $\uparrow$ & Acc. $\uparrow$ & F1 $\uparrow$ & MAE $\downarrow$ & RMSE $\downarrow$ & MAE $\downarrow$ & RMSE $\downarrow$ & R-1 $\uparrow$ & R-L $\uparrow$ & R-1 $\uparrow$ & R-L $\uparrow$ & R-1 $\uparrow$ & R-L $\uparrow$ & R-1 $\uparrow$ & R-L $\uparrow$ \\
    \midrule
    Zero-shot & 0.214 & 0.285 & 0.430 & 0.364 & 0.361 & 0.703 & 0.350 & 0.681 & 0.444 & 0.394 & 0.447 & 0.386 & 0.445 & 0.396 & 0.448 & 0.399 \\
    ICL & 0.279 & 0.336 & 0.495 & 0.415 & 0.333 & 0.662 & 0.322 & 0.641 & 0.452 & 0.395 & 0.461 & 0.405 & 0.453 & 0.397 & 0.462 & 0.418 \\
    RAG & 0.282 & 0.340 & 0.501 & 0.419 & 0.328 & 0.655 & 0.317 & 0.632 & 0.457 & 0.399 & 0.466 & 0.409 & 0.459 & 0.402 & 0.468 & 0.422 \\
    PAG & 0.287 & 0.345 & 0.509 & 0.426 & 0.323 & 0.646 & 0.312 & 0.623 & 0.463 & 0.404 & 0.473 & 0.415 & 0.465 & 0.406 & 0.475 & 0.428 \\
    HYDRA & 0.291 & 0.351 & 0.515 & 0.432 & 0.318 & 0.638 & 0.307 & 0.615 & 0.473 & 0.412 & 0.483 & 0.424 & 0.471 & 0.411 & 0.481 & 0.434 \\
    \midrule
    \textbf{RPO} & \textbf{0.355} & \textbf{0.400} & \textbf{0.571} & \textbf{0.479} & \textbf{0.252} & \textbf{0.564} & \textbf{0.241} & \textbf{0.540} & \textbf{0.498} & \textbf{0.425} & \textbf{0.507} & \textbf{0.435} & \textbf{0.499} & \textbf{0.427} & \textbf{0.508} & \textbf{0.448} \\
    \bottomrule
  \end{tabular}
\end{table*}

\begin{table*}[h!] 
\large
\centering
\caption{Ablation study of RPO components.}
\label{tab:ablation_results}
\begin{tabular}{lcccccccc}
\toprule
& \multicolumn{2}{c}{\textbf{LaMP-2}} & \multicolumn{2}{c}{\textbf{LaMP-3}} & \multicolumn{2}{c}{\textbf{LaMP-5}} & \multicolumn{2}{c}{\textbf{LaMP-7}} \\
\cmidrule(r){2-3} \cmidrule(r){4-5} \cmidrule(r){6-7} \cmidrule(l){8-9}
\textbf{Method} & \textbf{Acc. $\uparrow$} & \textbf{F1 $\uparrow$} & \textbf{MAE $\downarrow$} & \textbf{RMSE $\downarrow$} & \textbf{R-1 $\uparrow$} & \textbf{R-L $\uparrow$} & \textbf{R-1 $\uparrow$} & \textbf{R-L $\uparrow$} \\
Vanilla RAG & 0.213 & 0.162 & 0.332 & 0.635 & 0.382 & 0.354 & 0.487 & 0.415 \\
RAG w/ ICL & 0.256 & 0.212 & 0.329 & 0.633 & 0.393 & 0.362 & 0.490 & 0.418 \\
\midrule 
RPO w/o SFT & 0.342 & 0.313 & 0.324 & 0.628 & 0.415 & 0.379 & 0.496 & 0.425 \\
RPO w/o RL & 0.376 & 0.335 & 0.319 & 0.617 & 0.438 & 0.391 & 0.503 & 0.439 \\
\textbf{RPO (Full)} & \textbf{0.403} & \textbf{0.364} & \textbf{0.310} & \textbf{0.620} & \textbf{0.488} & \textbf{0.435} & \textbf{0.526} & \textbf{0.467} \\
\bottomrule
\end{tabular}
\end{table*}

\section{Experiments}
\subsection{Experimental Setup}
\noindent \textbf{Datasets and Tasks.} We evaluate our method on the LaMP benchm\-ark\cite{salemi2024optimization}, with dataset details provided in Appendix \ref{app:data}, a standard suite for testing LLM personalization across diverse user scenarios. We select four representative tasks: LaMP-2 (Movie Tagging), which predicts tags aligned with user-specific themes and styles; LaMP-3 (Product Rating), assessing sentiment in light of prior rating patterns; LaMP-5 (Scholarly Title Generation), requiring domain-aware and stylistically aligned title synthesis; and LaMP-7 (Tweet Paraphrasing), which tests the model’s ability to retain meaning while emulating user phrasing.

To mirror real-world personalization, we adopt two partitioning protocols from LaMP: (1) a chronological split preserving temporal user history across train/validation/test; and (2) a user-level split assigning 100 users for training and 50 disjoint users for testing.

\noindent \textbf{Baselines.} To evaluate the effectiveness of the proposed framework, we compare it against a comprehensive suite of personalization baselines. All methods leverage the frozen gpt-4.1-mini-2025-04-14 model and, where applicable, adopt a consistent retrieval setup: for each query, the top-4 user history items are retrieved using Contriever\cite{izacard2021unsupervised} and appended as in-context demonstrations. The use of gpt-4.1-mini-2025-04-14 as the core component ensures a fair comparison across all methods.

\textbf{(1) Zero-shot.} This fundamental baseline generates a response based solely on the task query, without any user history or personalized examples. It serves as a non-personalized lower bound to quantify the gains from personalization.

\textbf{(2) In-Context Learning (ICL).} This baseline prepends a fixed number of examples from the user's history to the task input, serving as in-context demonstrations. It provides a general personalization signal by leveraging historical context, without dynamically tailoring the examples to the current query.

\textbf{(3) Retrieval-Augmented Generation (RAG).} This baseline improves upon ICL by employing a dynamic, query-aware retrieval strategy. A pretrained Contriever model\cite{izacard2021unsupervised} retrieves user history items based on their semantic relevance to the current query. Following prior studies, the top-4 retrieved examples are prepended to the task input, providing a strong and targeted personalization signal. 

\textbf{(4) Profile-Augmented Generation (PAG).}\cite{richardson2023integrating} This method first synthesizes a textual user profile by summarizing the user's historical interactions and preferences. This generated profile is then prepended to the input prompt, providing the LLM with an explicit and condensed representation of the user's characteristics to guide generation.

\textbf{(5) HYDRA.} This baseline represents a state-of-the-art, learning-based personalization approach\cite{zhuang2024hydra}. HYDRA employs a retrieve-then-rerank strategy combined with a lightweight adapter to align the black-box LLM's outputs with user preferences. It is a powerful plug-and-play framework that significantly improves personalization without modifying the base model's parameters.

\noindent \textbf{Evaluation Metrics.} To ensure consistent and task-relevant assessment, we adopt the metric design prescribed by the LaMP benchmark, with tailored evaluation criteria for each selected task. For LaMP-2, which involves classification, we compute both accuracy and F1 score to capture performance on both overall correctness and label balance. LaMP-3 is framed as a regression problem, for which we report mean absolute error (MAE) and root mean squared error (RMSE)—two metrics that quantify deviation between predictions and ground-truth values, with lower scores indicating stronger predictive precision. For LaMP-5 and LaMP-7, which involve structured text generation, we employ ROUGE-1 and ROUGE-L to evaluate lexical overlap and sequence-level alignment with reference outputs. In these cases, higher ROUGE scores are indicative of better generative fidelity.

\noindent \textbf{Implementation Details.} We use Qwen2.5-7B-Instruct\cite{team2024qwen2} as the backbone of the reflection module, supporting sequences up to 2048 tokens to accommodate contextual cues and reasoning traces.

During SFT, the model is trained for one epoch with batch size 8, using AdamW\cite{loshchilov2017decoupled} (learning rate $5 \times 10^{-6}$, 5\% linear warm-up). We retrieve the top-1 relevant user example to simulate lightweight personalization within the prompt.

For RL, we conduct one training epoch with a progressive in-context strategy: 2–6 demonstrations are incrementally sampled per query to emulate varying user history density. The model generates 16 candidates via nucleus sampling (temperature 1.0, top-p 0.9). Actor and critic are updated separately with learning rates of $5 \times 10^{-7}$ and $9 \times 10^{-6}$.

\subsection{Main Results}
\label{sec:main}
The proposed RPO framework achieves state-of-the-art (SOTA) performance across all personalization tasks in the LaMP benchmark, consistently and significantly surpassing mainstream baseline methods. As summarized in Table~\ref{tab:lamp_results_progressive}, RPO yields substantial improvements over prior approaches across diverse evaluation metrics and task settings. Specifically, for the LaMP-2 user dimension task, RPO attains an Acc of 0.355 and an F1 score of 0.400, substantially outperforming the Zero-shot baseline (Acc: 0.214, F1: 0.285). On the LaMP-3 time dimension task, RPO reduces the MAE to 0.241 and the RMSE to 0.540, representing notable improvements compared to Zero-shot (MAE: 0.350, RMSE: 0.681). Furthermore, in the LaMP-5 user dimension task, RPO achieves an R-1 score of 0.498 and an R-L score of 0.425, outperforming the best baseline, HYDRA (R-1: 0.473, R-L: 0.412). RPO also demonstrates leading performance on the LaMP-7 task, achieving a user dimension R-1 of 0.499 and an R-L of 0.427.

RPO's core innovation lies in its "generate-and-refine" strategy. Instead of generating personalized content from scratch, it first leverages a base model to produce a semantically complete but generic initial response. This initial response then serves as a crucial contextual anchor, guiding a Reflection Module to retrieve and integrate significant personalized features from the user's history with higher precision. This approach effectively transforms an open-ended generation problem into a more constrained and goal-oriented rewriting task, thereby enhancing the accuracy and relevance of the generated content. To ensure the reliability and interpretability of the refinement process, RPO incorporates a structured Chain-of-Thought (CoT) reasoning trajectory. The entire framework achieves deeper user alignment than advanced adapters through a two-stage training pipeline. In the SFT stage, reliable personalization capabilities are injected through the reasoning trajectory; subsequently, the RL stage further optimizes the strategy to improve output quality and generalization performance. 

\subsection{Ablation Study}

To validate the contribution of each component of PRO, we decoupled its two-stage training framework and individually evaluated the gains from the SFT and RL stages.

We employed two different retrieval baselines. The first, Vanilla RAG, adds the retrieved user history to the input, with the base model (gpt-4.1-mini-2025-04-14) directly generating the final output. The second, a more robust baseline we term RAG w/ ICL, mimics PRO's "post-generation rewriting" architecture: after the base model generates a generic response, an external, untrained Qwen2.5-7B-Instruct model receives a prompt with few-shot examples to rewrite that response into a personalized version.

As shown in Table\ref{tab:ablation_results}, the experimental results reveal the decisive advantage of our training paradigm. The RPO w/o RL model, relying solely on the SFT stage, significantly outperformed the RAG w/ ICL baseline. In the LaMP-2 task, accuracy saw a substantial improvement from 0.256 to 0.376. A similarly stark contrast emerged in the LaMP-5 task, where its R-1 score climbed from 0.393 to 0.438. This strongly demonstrates that systematic supervised training, based on structured reasoning trajectories, can imbue the model with a deep and generalizable understanding of user preferences, an effect that prompt-driven in-context learning alone cannot replicate.

In contrast, directly using RL to train the reflection model (RPO w/o SFT) proved far less efficient. Although this model still surpassed the RAG w/ ICL baseline, its performance consistently remained significantly weaker than models initialized with SFT.

The culmination of this two-stage process is the complete RPO framework, which achieves optimal performance across all tasks through a powerful synergy between SFT and RL. The SFT stage introduces foundational reasoning and preference alignment capabilities, while the RL stage further refines the generation strategy through explorative updates. This synergy is crucial for enhancing the model's generalization ability and the quality of its final outputs when confronted with diverse user histories.

\subsection{Effectiveness of the Progressive Curriculum on Context Scaling}

\begin{figure}[!t]
  \includegraphics[width=\columnwidth]{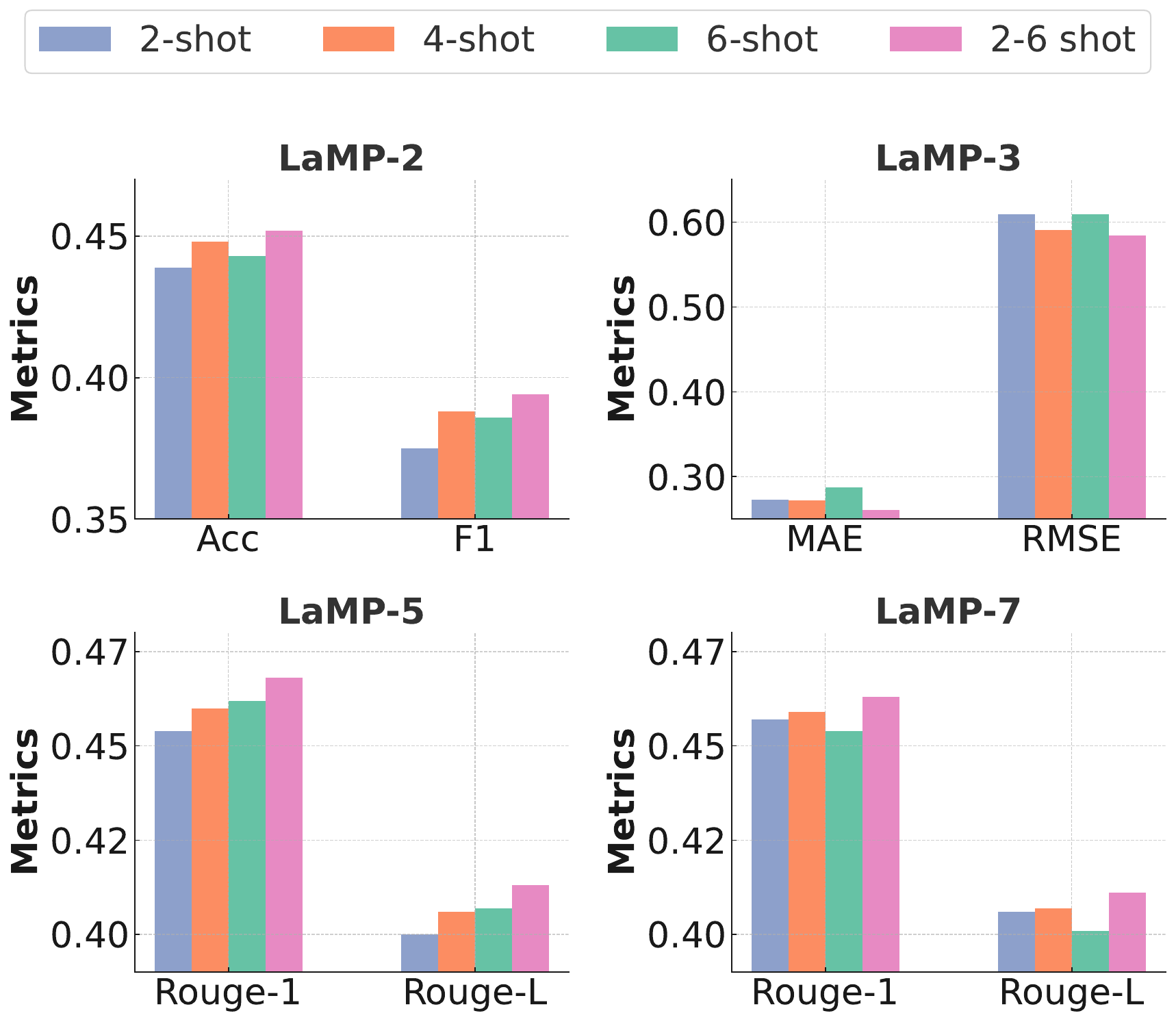}
  \caption{The figure illustrates how varying the number of shots influences RPO during the reinforcement learning phase. Results suggest that a progressive sampling strategy, which increases the shot count incrementally, leads to superior performance.}
  \label{fig:shot}
\end{figure}

To validate the effectiveness of our proposed Progressive Curriculum on context scaling, we conducted a comparative analysis against standard fixed-shot strategies. We systematically evaluated how the RPO framework's performance is influenced by the number of in-context examples (or "shots") drawn from user history.

Our experimental results (as shown in Figure \ref{fig:shot}) first establish a baseline using fixed-shot settings. Specifically, when the number of in-context examples increased from 2-shot to 4-shot, the model showed significant gains across multiple tasks. For example, in the LaMP-2 task, both Acc and F1 score improved. This suggests that a moderate amount of historical information is crucial for the model to capture users' core preferences.

However, these fixed-shot approaches reveal a critical limitation. When the number of examples further increased to 6-shot, we observed that performance stabilized or even experienced a slight decline. This phenomenon indicates that excessive contextual information can introduce noise, especially when less relevant historical records are included. Such records dilute or interfere with the model's judgment of the user's core style, thereby negatively impacting the final personalization effect.

This is precisely where the effectiveness of our progressive curriculum becomes evident. The strategy, represented as "2-6 shot" in the figure, achieved the best or near-best results on almost all tasks by outperforming all fixed-shot configurations. This empirically validates our central hypothesis: a curriculum-based approach to context scaling is superior to any fixed-shot setting. It confirms that by first learning from clean, focused signals and then generalizing to more complex contexts, RPO can develop a more robust and effective personalization policy, striking an optimal balance between leveraging rich history and avoiding noise.

\begin{table}[t]
\centering
\normalsize
\setlength{\tabcolsep}{5pt}
\renewcommand{\arraystretch}{1.15}
\caption{The effect of different retriever models on the performance of RPO.}
\begin{tabular}{llccc}
\toprule
\textbf{Task} & \textbf{Metrics} & \textbf{Random} & \textbf{Contriever} & \textbf{BGE} \\
\midrule
\multirow{2}{*}{LaMP-2} & Accuracy $\uparrow$  & 0.535 & \textbf{0.565} & 0.555 \\
                        & F1 $\uparrow$   & 0.450 & \textbf{0.501} & 0.500 \\
\multirow{2}{*}{LaMP-3} & MAE $\downarrow$  & 0.351 & \textbf{0.231} & 0.233 \\
                        & RMSE $\downarrow$ & 0.747 & 0.534 & \textbf{0.528} \\
\multirow{2}{*}{LaMP-5} & Rouge-1 $\uparrow$  & 0.437 & \textbf{0.487} & 0.483 \\
                        & Rouge-L $\uparrow$  & 0.407 & 0.436 & \textbf{0.437} \\
\multirow{2}{*}{LaMP-7} & Rouge-1 $\uparrow$  & 0.486 & 0.536 & \textbf{0.537} \\
                        & Rouge-L $\uparrow$  & 0.455 & \textbf{0.484} & 0.481 \\
\bottomrule
\end{tabular}
\label{tab:retriever}
\end{table}

\subsection{Analysis of the Retriever Model}

To assess the importance of the retriever module as a cornerstone of the RPO framework's performance, we systematically analyzed the framework's performance when equipped with different retrievers: a random baseline and two advanced semantic retrieval models, Contriever and BGE\cite{xiao2024c}.

Our experimental results clearly demonstrate that high-quality semantic retrieval is a prerequisite for effective personalization. As shown in Table \ref{tab:retriever}, using either Contriever or BGE leads to overwhelming performance improvements across all tasks compared to randomly selecting user history. For instance, in the LaMP-3 task, BGE significantly reduced the MAE from the random baseline's 0.351 to 0.231.

In a head-to-head comparison between the two advanced retrievers, we found their performance to be quite comparable, with only minor, non-systematic differences across various metrics. Specifically, on the LaMP-2 classification task, Contriever achieved the highest scores for both Acc (0.565) and F1 score (0.501). Conversely, on the LaMP-7 generation task, BGE slightly outperformed Contriever in the ROUGE-1 score (0.537 vs. 0.536). This minimal performance difference indicates that the RPO framework is not sensitive to the specific choice of advanced retriever, but rather can robustly collaborate with any high-quality semantic retrieval model.

\subsection{Analysis of CoT Methods}

In this experimental section, we explored the differing roles of CoT in two paradigms: first, the "direct generation" mode, where CoT is executed via in-context prompting within traditional baseline methods; and second, the "post-processing" mode, where CoT optimizes answers through rewriting within the RPO framework.

To ensure a fair comparison, we extended three types of baseline methods using the CoT method. This extension enabled them to explicitly execute CoT through in-context prompting before generating answers, thereby simulating the user's preferred reasoning process. The experimental results are presented in Table \ref{tab:deepseek_with_pag_varied}.

Our experimental results clearly demonstrate that the RPO framework significantly outperforms the CoT-enhanced direct generation baselines across all tasks. For instance, in the LaMP-5 task, RPO achieved a ROUGE-1 score of 0.489, representing a notable improvement over the best-performing baseline method (0.465). This outcome indicates that, with the base model generating initial content, RPO's CoT process can focus on optimizing style and phrasing. Its two-stage learning mechanism aligns well with the ability to finely capture user historical features in various personalization tasks. In contrast, CoT in the direct generation mode must simultaneously handle content generation and style alignment, leading to an unclear division of labor and consequently limiting performance. Similarly, in the LaMP-3 task, RPO reduced the Mean Absolute Error (MAE) to 0.283, comprehensively surpassing all baseline methods. This result further validates that reflecting on and revising existing ratings is more efficient and accurate than reasoning from scratch.

\begin{table}[t]
\centering
\normalsize
\setlength{\tabcolsep}{5pt}
\renewcommand{\arraystretch}{1.15}
\caption{Performance comparison of CoT-enhanced baselines versus the RPO framework.}
\begin{tabular}{l l c c c c}
\toprule
\multirow{2}{*}{\textbf{Task}} & \multirow{2}{*}{\textbf{Metrics}} & \multicolumn{1}{c}{Zero-shot} & \multicolumn{1}{c}{RAG} & \multicolumn{1}{c}{PAG} & \multirow{2}{*}{\textbf{RPO}} \\
\cline{3-5}
& & \multicolumn{1}{c}{w/ CoT} & \multicolumn{1}{c}{w/ CoT} & \multicolumn{1}{c}{w/ CoT} & \\
\midrule
\multirow{2}{*}{LaMP-2} 
& Accuracy $\uparrow$ & 0.429 & 0.437 & 0.443 & \textbf{0.456} \\
& F1 $\uparrow$ & 0.354 & 0.375 & 0.380 & \textbf{0.393} \\
\midrule
\multirow{2}{*}{LaMP-3} 
& MAE $\downarrow$ & 0.457 & 0.354 & 0.341 & \textbf{0.283} \\
& RMSE $\downarrow$ & 0.743 & 0.651 & 0.626 & \textbf{0.597} \\
\midrule
\multirow{2}{*}{LaMP-5} 
& Rouge-1 $\uparrow$ & 0.436 & 0.454 & 0.465 & \textbf{0.489} \\
& Rouge-L $\uparrow$ & 0.375 & 0.395 & 0.405 & \textbf{0.437} \\
\midrule
\multirow{2}{*}{LaMP-7} 
& Rouge-1 $\uparrow$ & 0.464 & 0.448 & 0.462 & \textbf{0.539} \\
& Rouge-L $\uparrow$ & 0.405 & 0.401 & 0.413 & \textbf{0.482} \\
\bottomrule
\end{tabular}
\label{tab:deepseek_with_pag_varied}
\end{table}

\subsection{Analysis of Insensitivity to the Base Model}
\label{sec:base_model_insensitivity}

To validate one of the core advantages of the RPO framework---its model-agnostic nature---we conducted a crucial experiment. In this setup, we kept the trained reflection model constant while swapping the black-box base model responsible for generating the initial, non-personalized response. We selected three leading and distinct LLMs as the base model: DeepSeek-V3-0324 (DeepSeek-V3), Qwen3-30B-A3B-2507 (Qwen3), and gpt-4o-mini-2024-07-18 (GPT-4o-mini).

The experimental results demonstrate a high degree of consistency. Regardless of the base model used, the final personalization performance of the RPO framework remained exceptionally high and stable. As shown in Table~\ref{tab:metrics_comparison_randomized}, on the LaMP-2 classification task, the accuracy scores were 0.465, 0.469, and 0.467 respectively, with F1 scores showing similarly minimal fluctuation (0.390, 0.394, and 0.392). This stability was also evident in the LaMP-3 regression task for MAE and RMSE metrics, as well as in the ROUGE-1 and ROUGE-L scores for generation tasks like LaMP-7.

These findings strongly confirm that RPO's decoupled design is successful. They indicate that the final quality of the personalized output is not heavily dependent on the unique style or inherent biases of a particular base model. As long as the base model can produce a semantically accurate and high-quality generic response, RPO's reflection module can consistently and effectively perform its rewriting and alignment tasks.

This characteristic holds significant practical implications. It means RPO can serve as a plug-and-play personalization layer, flexibly integrating with any existing or future black-box LLM API without requiring retraining of the reflection module. Users can freely choose the most suitable base model based on factors like cost, inference speed, or availability without sacrificing the quality of the final personalization. This fully embodies RPO's potential as an efficient, modular, and truly model-agnostic solution.

\begin{table}[t]
    \centering
    \normalsize
    \setlength{\tabcolsep}{2.5pt} 
    \renewcommand{\arraystretch}{1.15} 
    \caption{Performance comparison of RPO across various competitive base models.}
    \label{tab:metrics_comparison_randomized}
    \begin{tabular}{llccc}
        \toprule
        \textbf{Task} & \textbf{Metrics} & \textbf{DeepSeek-V3} & \textbf{Qwen3} & \textbf{GPT-4o-mini} \\
        \midrule
        \multirow{2}{*}{LaMP-2} & Accuracy $\uparrow$ & 0.465 & \textbf{0.469} & 0.467 \\
                                & F1 $\uparrow$       & 0.390 & \textbf{0.394} & 0.392 \\
        \midrule
        \multirow{2}{*}{LaMP-3} & MAE $\downarrow$    & \textbf{0.281} & 0.285 & 0.283 \\
                                & RMSE $\downarrow$   & \textbf{0.596} & 0.601 & 0.599 \\
        \midrule
        \multirow{2}{*}{LaMP-5} & Rouge-1 $\uparrow$  & 0.475 & 0.472 & \textbf{0.477} \\
                                & Rouge-L $\uparrow$  & 0.414 & 0.411 & \textbf{0.416} \\
        \midrule
        \multirow{2}{*}{LaMP-7} & Rouge-1 $\uparrow$  & \textbf{0.551} & 0.547 & 0.549 \\
                                & Rouge-L $\uparrow$  & \textbf{0.490} & 0.485 & 0.488 \\
        \bottomrule
    \end{tabular}
\end{table}

\section{Conclusion}

We propose RPO, a novel framework that reconceptualizes large language model (LLM) personalization as a transparent, reasoning-driven post-hoc optimization problem, fundamentally departing from traditional black-box conditioning approaches. At the core of RPO lies a lightweight reflection module, which learns to emulate user-specific reasoning policies through a two-stage process: first, supervised fine-tuning on structured Rewriting Trajectories to capture user preferences, and second, reinforcement learning with a progressive multi-context curriculum to enhance generalization across diverse interaction histories. Extensive empirical evaluations demonstrate that RPO achieves state-of-the-art performance across a wide range of classification, regression, and generation tasks, consistently outperforming competitive baselines. While the current implementation employs a compact 7B reflection module to ensure deployment efficiency, the framework itself is base-model agnostic and holds potential for further improvements when scaled to more powerful reasoning models. By modeling user-specific reasoning, RPO establishes a modular, interpretable, and scalable paradigm for LLM personalization, bridging the gap between implicit context injection and explicit response shaping.



\bibliographystyle{ACM-Reference-Format}
\bibliography{custom}

\clearpage
\newpage
\appendix

\section{Dataset Details}
\label{app:data}
\paragraph{LaMP Benchmark Overview.}  
The LaMP (Language Model Personalization) benchmark is a standardized evaluation suite designed to assess how effectively large language models (LLMs) can adapt to user-specific preferences, linguistic styles, and behavioral traits. LaMP provides a diverse set of personalization tasks spanning classification, regression, and generation, with structured user profiles and chronologically ordered histories. Each user is associated with a series of task-specific records, allowing both personalized modeling and robust evaluation of generalization across unseen users or future interactions.

LaMP adopts two canonical evaluation settings:  
\begin{itemize}
    \item \textbf{User Split:} models are trained on interactions from a subset of users (100) and evaluated on disjoint users (50). This tests user-level generalization.
    \item \textbf{Time Split:} models are trained on the early histories of users and evaluated on their future interactions. This tests temporal adaptation.
\end{itemize}

\begin{table*}[t]
\centering
\footnotesize
\caption{LaMP task statistics under user/time separation.}
\label{tab:lamp_stats_selected}
\renewcommand{\arraystretch}{1.15}
\setlength{\tabcolsep}{3pt}
\begin{tabular}{lllcccccc}
\toprule
\textbf{Task} & \textbf{Type} & \textbf{Separation} & \textbf{\#Train} & \textbf{\#Dev} & \textbf{Input Len.} & \textbf{Output Len.} & \textbf{Profile Size} & \textbf{\#Classes} \\
\midrule
\multirow{2}{*}{LaMP-2} & \multirow{2}{*}{binary classification} 
 & user & 9682 & 2500 & 51.40 ± 5.72 & - & 90.61 ± 53.87 & 2 \\
 &  & time & 6542 & 1500 & 51.43 ± 5.70 & - & 84.15 ± 47.54 & 2 \\
\midrule
\multirow{2}{*}{LaMP-3} & \multirow{2}{*}{categorical classification} 
 & user & 3820 & 692 & 92.27 ± 20.83 & - & 159.29 ± 330.81 & 15 \\
 &  & time & 5073 & 1410 & 92.39 ± 21.95 & - & 86.76 ± 189.52 & 15 \\
\midrule
\multirow{2}{*}{LaMP-5} & \multirow{2}{*}{text generation} 
 & user & 9682 & 2500 & 152.81 ± 86.60 & 9.26 ± 3.13 & 89.61 ± 53.87 & - \\
 &  & time & 14682 & 1500 & 162.34 ± 65.63 & 9.71 ± 3.21 & 87.88 ± 53.63 & - \\
\midrule
\multirow{2}{*}{LaMP-7} & \multirow{2}{*}{text generation} 
 & user & 10437 & 1500 & 29.16 ± 6.94 & 16.93 ± 6.55 & 17.74 ± 15.10 & - \\
 &  & time & 13437 & 1498 & 29.72 ± 7.01 & 16.96 ± 5.67 & 15.71 ± 14.86 & - \\
\bottomrule
\end{tabular}
\end{table*}

\noindent
Across all tasks, the dataset comprises tens of thousands of user– interaction records, with rich textual inputs and profile features. Table \ref{tab:lamp_stats_selected} summarizes key statistics for the four LaMP tasks used in our work, including data split sizes, average input/output lengths, and the number of user classes or tags. For each task, both user and time separation strategies are provided to facilitate cross-user and temporal evaluation.

\paragraph{Tasks Used in This Work.}  
In our experiments, we focus on four representative LaMP tasks that differ in output type, task complexity, and personalization requirements:

\begin{itemize}
  \item \textbf{LaMP-2 (Movie Tagging)}: A multi-label classification task where the goal is to predict movie-related tags (e.g., genre, theme, tone) that align with the user's labeling behavior.
  \item \textbf{LaMP-3 (Product Rating)}: A regression task in which the model predicts a user’s rating (on a 1–5 scale) for a product.
  \item \textbf{LaMP-5 (Scholarly Title Generation)}: A text generation task where the model must rewrite academic paper titles to reflect a user’s stylistic preferences or domain-specific phrasing.
  \item \textbf{LaMP-7 (Tweet Paraphrasing)}: A stylistic paraphrasing task requiring the model to rephrase a given tweet in the user's unique tone or writing style.
\end{itemize}

\section{Prompt}
\label{app:prompt}
To enable consistent training and evaluation across heterogeneous personalization tasks in the LaMP benchmark, we define structured prompt templates tailored to each task’s input-output format, as showen in Table \ref{tab:prompt_templates}. These prompts serve two key purposes: (1) they guide the base model in generating generic outputs; and (2) they form the foundation for constructing personalized reasoning trajectories used to supervise the reflection module. While differing in surface form, all prompts share a minimalistic and instruction-oriented structure, encouraging models to focus on the semantic core of the task without introducing extraneous formatting or verbosity. Below, we present the unified prompt template encompassing the four representative LaMP tasks used in our study: LaMP-2 (movie tagging), LaMP-3 (product rating), LaMP-5 (title generation), and LaMP-7 (tweet paraphrasing).

These prompts are intentionally kept concise to minimize instruction leakage and maximize compatibility with reasoning-based reflection. For classification (LaMP-2) and regression (LaMP-3), prompts impose strict formatting constraints to enable numeric or categorical evaluation. For generation tasks (LaMP-5 and LaMP-7), prompts are open-ended but avoid stylistic cues, deferring all personalization decisions to the reflection module. In all cases, only the task input is visible to the base model; personalized information is introduced post-hoc through the RPO pipeline. This design enforces a clear separation between generic and user-aligned output spaces, ensuring the reflection module learns to inject personalization in a controlled, interpretable manner.

\begin{table*}[t]
\centering
\small
\renewcommand{\arraystretch}{1.2}
\setlength{\tabcolsep}{3pt}
\caption{Unified prompt templates used for the four representative LaMP tasks. Prompts are intentionally concise to enable consistent base model outputs and serve as the foundation for reflective personalization.}
\label{tab:prompt_templates}
\begin{tabular}{@{}llp{11.5cm}@{}}
\toprule
\textbf{Task} & \textbf{Type} & \textbf{Prompt Template} \\
\midrule
LaMP-2 & Movie Tagging (Classification) &
\texttt{Which tag does this movie relate to among the following tags?}\\
& & \texttt{Just answer with the tag name without further explanation.}\\
& & \texttt{tags: [sci-fi, based on a book, comedy, action, twist ending, dystopia, dark comedy, classic, ...]}\\
& & \texttt{description: [MOVIE]} \\

\addlinespace[0.5em]

LaMP-3 & Product Rating (Regression) &
\texttt{What is the score of the following review on a scale of 1 to 5?}\\
& & \texttt{Just answer with 1, 2, 3, 4, or 5 without further explanation.} \\
& & \texttt{review: [REVIEW]} \\

\addlinespace[0.5em]

LaMP-5 & Title Generation (Generation) &
\texttt{Generate a title for the following abstract of a paper:}\\
& & \texttt{abstract: [ABSTRACT]} \\

\addlinespace[0.5em]

LaMP-7 & Tweet Paraphrasing (Generation) &
\texttt{Paraphrase the following tweet without any explanation before or after it:}\\
& & \texttt{[TWEET]} \\
\bottomrule
\end{tabular}
\end{table*}

\section{Case Study}
\label{app:case}
To complement the quantitative results and offer a more nuanced understanding of how the proposed RPO framework operates in practice, we now turn to a series of illustrative case studies drawn from the LaMP benchmark. While aggregated metrics such as accuracy, ROUGE, and MAE offer broad insight into model effectiveness, they often fail to capture the subtle ways in which personalization manifests at the level of individual predictions. In contrast, case studies enable a fine-grained examination of how the reflection module adapts generic outputs to align with a user’s unique linguistic habits, semantic priorities, and interpretive patterns.

Each case presented below includes a user query, a base response generated without personalization, and the final personalized output produced by the RPO model. Accompanying these examples is a brief interpretive commentary (<think>) that reflects the reasoning trajectory employed by the model to reconcile the base output with stylistic cues and contextual signals extracted from the retrieved user history. This structured format is designed to reveal not only the final outcome of personalization but also the underlying transformation logic that drives the shift from generality to specificity.

We select representative instances from three distinct task types— classification (LaMP-2), regression (LaMP-3), and generation (LaMP-5 and LaMP-7)—to demonstrate RPO’s versatility across diverse output modalities. Through these concrete examples, we aim to showcase how RPO supports interpretable, controllable, and consistent user alignment, even in the presence of sparse, noisy, or stylistically heterogeneous profiles.

\begin{figure*}[t]
\centering
\begin{tcolorbox}[
  colback=gray!3,
  title=\textbf{RPO Prompt Demo for LaMP-2},
  fonttitle=\bfseries\small,
  width=\textwidth,
  boxrule=0.4pt,
  arc=2pt,
  outer arc=2pt,
  breakable,
  enhanced
]
\ttfamily
\footnotesize
\noindent
\textbf{Prompt:} You will receive a user's query, an initial response, and a set of the user's past movie description–tag pairs as examples of their tagging preferences and interpretive patterns. Analyze these prior examples to infer the user’s personalized tagging style, including how they interpret thematic elements, prioritize certain genres or motifs, and distinguish subtle differences between similar tags. Then, revise the initial tag prediction to better reflect this personalized tagging behavior. Do not copy tag assignments directly from the examples—only emulate the user’s decision-making tendencies. The output should follow this structured format: <think>...</think>, and <personalized>...</personalized>.

\medskip

\textbf{User profile:} \\
The tag for the movie: "Former Special Forces officer, Frank Martin will deliver anything to anyone for the right price, and his no-questions-asked policy puts him in high demand. But when he realizes his latest cargo is alive, it sets in motion a dangerous chain of events..." is \texttt{action} \\
The tag for the movie: "A plastic surgeon, romancing a much younger schoolteacher, enlists his loyal assistant to pretend to be his soon to be ex-wife, in order to cover up a careless lie..." is \texttt{comedy} \\
The tag for the movie: "Frank Martin puts the driving gloves on to deliver Valentina, the kidnapped daughter of a Ukrainian official, from Marseilles to Odessa..." is \texttt{action} \\
The tag for the movie: "Viktor Navorski is a man without a country... now he's stranded at Kennedy Airport, where he's holding a passport that nobody recognizes..." is \texttt{comedy}

\medskip

\textbf{User query:} Which tag does this movie relate to among the following tags? Just answer with the tag name without further explanation. \\
\textbf{Tags:} [sci-fi, based on a book, comedy, action, twist ending, dystopia, dark comedy, classic, psychology, fantasy, romance, thought-provoking, social commentary, violence, true story] \\
\textbf{Description:} Overwhelmed by her suffocating schedule, touring European princess Ann takes off for a night while in Rome. When a sedative she took from her doctor kicks in, she falls asleep on a park bench and is found by an American reporter...

\textbf{Initial response:} \texttt{romance}

\medskip

\textbf{Model output:} \\
<think> The user's tagging history shows a pattern where they use 'action' for high-stakes missions, rescues, and physical conflict, often involving professionals like Special Forces officers. For 'comedy', they prioritize situational humor, misunderstandings, and chaotic relationships (e.g., fabricated ex-wives, accidental airport living). Notably, romance is present in their comedy examples but isn’t the primary tag driver—instead, the focus is on absurdity and contrived scenarios. The current movie description centers on a reporter exploiting a princess’s incognito night out for a story, creating a lie-driven premise with humorous potential (sedative mishap, secret identity). While romance develops, the setup mirrors the user’s comedy precedents (ulterior motives, escalating deceptions) more than pure romantic focus. </think> \\
<personalized> comedy </personalized>
\end{tcolorbox}
\label{case:lamp2}
\end{figure*}

\begin{figure*}[t]
\centering
\begin{tcolorbox}[
  colback=gray!3,
  title=\textbf{RPO Prompt Demo for LaMP-3},
  fonttitle=\bfseries\small,
  width=\textwidth,
  boxrule=0.4pt,
  arc=2pt,
  outer arc=2pt,
  breakable,
  enhanced
]
\ttfamily
\footnotesize
\noindent
\textbf{Prompt:} You will receive a user's query, an initial response, and a set of the user's past review-score pairs as examples of their scoring behavior and stylistic tendencies. Analyze these prior examples to infer the user’s personalized rating style, including their implicit rating criteria, consistency, leniency or harshness in scoring, and any contextual biases. Then, revise the initial score to better match this personalized scoring behavior. Do not copy wording from prior examples—only emulate the underlying judgment style and decision logic. The output should follow this structured format: <think>...</think>, and <personalized>...</personalized>.

\medskip

\textbf{User profile:} \\
\texttt{3} is the score for \textit{"Works OK but I agree with the other reviewers--seems flimsy, the lights are way too bright, it only has one live connector into a socket (the other is 3 small plastic pieces which go into the bottom socket) which makes it a bit unsteady. I'm using it because I haven't seen anything much better."} \\
\texttt{2} is the score for \textit{"Works just as it's supposed to! Very useful in my long, dark closet. Will buy more. \\
UPDATE Used this about once a day since I purchased in November 2017. Today the switch stopped working--just rocks back and forth without making any contact. Since this was under \$10 I guess I should have expected it, but I had a similar one that worked for more than 8 years before going bad and didn't cost much more than this. i guess I'll try another brand this time."} \\
\texttt{4} is the score for \textit{"Nice product, but missing the stakes for securing to ground--I just used some rods I had in the garage, which worked fine. Very easy to put together, but follow the instruction carefully. If you get something wrong and need to switch a panel or end bracket, it's rather difficult to remove it to make the correction. Overall, I'm pleased with this purchase and may order another."} \\
\texttt{3} is the score for \textit{"This is a nice little unit. I got it because I had purchased a RING wireless video doorbell and my wi-fi signal was not strong enough to work on my front porch. The unit was a bit difficult to set up but I managed to do it using tips from other amazon reviewers. However, while it connected easily to my PC and RING, it simply would not connect to my apple devices (even with tech support help). Rather than replace it, I returned it and for \$45 more I upgraded to a new wireless router with a significantly stronger signal than my older one. That was well worth the extra \$\$, as I now have only one piece of equipment to deal with, not two; I have a great signal everywhere in my home; and I'm using one fewer electrical outlet. Not the right solution for everyone, but it was right for my needs."}

\medskip

\textbf{User query:} What is the score of the following review on a scale of 1 to 5? Just answer with 1, 2, 3, 4, or 5 without further explanation. \\
\textbf{Review:} \textit{"One was broken and unusable, another was cracked but I'm using it."} \\
\textbf{Initial response:} \texttt{2}

\medskip

\textbf{Model output:} \\
<think> The user's past reviews suggest they weigh functionality and durability heavily. When products fail prematurely (as in the 2-star case), they score harshly. However, if the product is usable despite flaws (3-star cases), they show leniency. The current review mentions partial usability (one broken, one cracked but used). The initial score of 2 aligns with the 2-star example where a product failed completely. However, since part of the product is still functional here, the user's pattern of giving 3 stars for "works but flawed" (like the flimsy lights and WiFi extender cases) suggests a 3 would better match their tendency to acknowledge partial utility. </think> \\
<personalized>3</personalized>
\end{tcolorbox}
\label{case:lamp3}
\end{figure*}

\begin{figure*}[t]
\centering
\begin{tcolorbox}[
  colback=gray!3,
  title=\textbf{RPO Prompt Demo for LaMP-5},
  fonttitle=\bfseries\small,
  width=\textwidth,
  boxrule=0.4pt,
  arc=2pt,
  outer arc=2pt,
  breakable,
  enhanced
]
\ttfamily
\footnotesize
\noindent
\textbf{Prompt:} You will receive a user's query, an initial response, and several abstract–title pairs as few-shot examples. Analyze the abstract-to-title transformation patterns demonstrated in the user's prior examples, focusing on their personalized titling style. Your task is to revise the initial title to align with this personalized style. Do not copy content—focus only on stylistic transformation. The output should follow this structured format: <think>...</think>, and <personalized>...</personalized>.

\medskip

\textbf{User profile:} \\
\texttt{"Very Sparse Stable Random Projections, Estimators and Tail Bounds for Stable Random Projections"} is the title for an abstract discussing methods to reduce cost using sparse stable random projections and analyzing their convergence rates. \\
\texttt{"Estimators and tail bounds for dimension reduction in $l_1$ ($0 < \alpha \le 2$) using stable random projections." } is the title for an abstract proposing new estimators and complexity bounds for stable projections with small $\alpha$. \\
\texttt{"A Very Efficient Scheme for Estimating Entropy of Data Streams Using Compressed Counting"} is the title for an abstract introducing Compressed Counting to improve entropy estimation as $\alpha \to 1$. \\
\texttt{"Sparse Recovery with Very Sparse Compressed Counting."} is the title for an abstract applying sparse compressed counting to signal recovery under skewed $p$-stable distributions.

\medskip

\textbf{User query:} Following the given patterns. Generate a title for the following abstract of a paper: \\
\textit{"We propose skewed stable random projections for approximating the $\alpha$th frequency moments of dynamic data streams ($0 < \alpha < 2$). We show the sample complexity $k = G \cdot \frac{1}{\epsilon^2} \log\left(\frac{2}{\delta}\right)$, where $G = O(\epsilon)$ as $\alpha \to 1$, improving prior bounds that required constant $G$ even as $\alpha \to 1$. This regime is practically relevant due to use cases involving small $\alpha$. Our method supports general Turnstile models with insertions/deletions, and defines the $\alpha$-moment $F^{(\alpha)} = \sum_{i=1}^{D} |A_t(i)|^\alpha$ for streaming signals $A_t$. The approach functions as a generalized counter accommodating decay/interest effects in dynamic contexts."} \\
\textbf{Initial response:} \texttt{Skewed Stable Random Projections for Approximating Frequency Moments in Dynamic Data Streams}

\medskip

\textbf{Model output:} \\
<think> The user's prior titles emphasize specific methodological innovations and their benefits, often using compound structures with commas or colons. For example, "Very Sparse Stable Random Projections, Estimators and Tail Bounds..." combines three focal points, while "A Very Efficient Scheme..." highlights efficiency. The abstract here introduces \textit{skewed stable random projections} with a key improvement in sample complexity ($G = O(\epsilon)$ as $\alpha \to 1$), addressing limitations of symmetric projections. The initial title ("Skewed Stable Random Projections for Approximating Frequency Moments...") captures the core method and application but lacks explicit reference to the \textit{improved sample complexity} or the \textit{$\alpha \to 1$ regime}, which are central to the technical contribution. \\
User titles often integrate both the method and its quantitative advantages (e.g., "Sparse Recovery with Very Sparse Compressed Counting" pairs sparsity with application). To align, the revised title should emphasize the efficiency gain in the $\alpha \to 1$ regime while retaining the dynamic data stream context. Using a compound structure with "Improved Sample Complexity" and "as $\alpha \to 1$" mirrors the user’s pattern of pairing technical scope with performance enhancements. </think> \\
<personalized>Skewed Stable Random Projections, Improved Sample Complexity for $\alpha$th Frequency Moments in Dynamic Data Streams as $\alpha \to 1$</personalized>
\end{tcolorbox}
\label{case:lamp5}
\end{figure*}

\begin{figure*}[t]
\centering
\begin{tcolorbox}[
  colback=gray!3,
  title=\textbf{RPO Prompt Demo for LaMP-7},
  fonttitle=\bfseries\small,
  width=\textwidth,
  boxrule=0.4pt,
  arc=2pt,
  outer arc=2pt
]
\ttfamily
\footnotesize
\noindent
\textbf{Prompt:} You will receive a user's query, an initial response, and a set of the user's past tweet replies as stylistic examples. Analyze the user's prior comments to infer their personalized paraphrasing style, including tone, phrasing, sentence structure, and rhetorical choices. Then, revise the initial paraphrase to match this personalized style. Do not copy wording from prior examples—only emulate their stylistic patterns. The output should follow this structured format: <think>...</think>, and <personalized>...</personalized>.

\medskip

\textbf{User profile:} \\
\textit{"I'm tired, been up all night.. its almost 5:20am, can't sleep yet though, got things to do!"} \\
\textit{"been awake for almost 24 hours, but off to london sooonish"} \\
\textit{"- haha i spent too much time in the sun! my rather sunburnt face"} \\
\textit{"i should be out sunbathing... hmm.. i might aswell before i go out"}

\medskip

\textbf{User query:} Following the given patterns. Paraphrase the following tweet without any explanation before or after it: \\
\textit{"Although I have a lot to do today, I am currently feeling unoccupied."} \\
\textbf{Initial response:} \texttt{"I have a lot to do today, but I am currently feeling unoccupied."}

\medskip

\textbf{Model output:} \\
<think> The user's past tweets exhibit a casual, conversational tone with frequent use of contractions ("I'm", "can't", "it's"), informal phrasing ("sooonish", "haha", "aswell"), and ellipses/em-dashes for pacing. They often juxtapose obligations with current states using "but" or "though" (e.g., "tired... can't sleep yet though"). The structure tends to start with a situation, then introduce a contrasting action/state. Time references ("5:20am", "24 hours") and physical sensations ("sunburnt face") are common. The personalized version should mirror this by: 1) Using contractions 2) Adding temporal context 3) Creating a "busy vs unoccupied" contrast with "but/though" 4) Employing ellipses for conversational flow. </think> \\
<personalized>Got a ton to do today… but here I am just sitting around right now though</personalized>
\end{tcolorbox}
\label{case:lamp7}
\end{figure*}

\end{document}